\documentclass[journal,draftcls,onecolumn]{IEEEtran} 

\usepackage{SIunits}
\usepackage[ugly]{units}
\usepackage[utf8x]{inputenc}
\usepackage[T1]{fontenc}
\usepackage{multirow}
\usepackage{amsmath,epsfig}
\usepackage{bm,bbm,booktabs}
\usepackage{subfigure}
\usepackage{float,rotating}
\usepackage{color}
\usepackage{cite}

\newcommand\hphi{\mbox{$h\mbox{-}\phi$}}

\title{Classification of Segments in PolSAR Imagery by Minimum Stochastic Distances Between Wishart Distributions}

\author{Wagner B.\ Silva, Corina C.\ Freitas, Sidnei J.\ S.\ Sant'Anna,~\IEEEmembership{Member} and Alejandro C.\ Frery,~\IEEEmembership{Member}}

\begin{document}

\maketitle
\markboth{Journal of Selected Topics in Applied Earth Observations and Remote Sensing,~Vol.~6, No.~3, June~2013}%
{Silva \MakeLowercase{\textit{et al.}}}
\begin{abstract}
A new classifier for Polarimetric SAR (PolSAR) images is proposed and assessed in this paper.
Its input consists of segments, and each one is assigned the class which minimizes a stochastic distance.
Assuming the complex Wishart model, several stochastic distances are obtained from the $\hphi$ family of divergences, and they are employed to derive hypothesis test statistics that are also used in the classification process. 
This article also presents, as a novelty, analytic expressions for the test statistics based on the following stochastic distances between complex Wishart models: Kullback-Leibler, Bhattacharyya, Hellinger, Rényi, and Chi-Square; also, the test statistic based on the Bhattacharyya distance between multivariate Gaussian distributions is presented.
The classifier performance is evaluated using simulated and real PolSAR data.
The simulated data are based on the complex Wishart model, aiming at the analysis of the proposal well controlled data. 
The real data refer to the complex L-band image, acquired during the 1994 SIR-C mission.
The results of the proposed classifier are compared with those obtained by a Wishart per-pixel/contextual classifier, and we show the better performance of the region-based classification. 
The influence of the statistical modeling is assessed by comparing the results using the Bhattacharyya distance between multivariate Gaussian distributions for amplitude data. 
The results with simulated data indicate that the proposed classification method has a very good performance when the data follow the Wishart model. 
The proposed classifier also performs better than the per-pixel/contextual classifier and the Bhattacharyya Gaussian distance using SIR-C PolSAR data.
\end{abstract}


\section{Introduction}
\label{sec:intro}

\IEEEPARstart{T}{he} classification of images obtained by polarimetric synthetic aperture radar (PolSAR) sensors is one of the main information extraction techniques from that kind of data. 
Generally, PolSAR classification falls into three categories: target decomposition~\cite{leepottier08},  PolSAR data statistical modeling~\cite{4305361}, and hybrid methods~\cite{789621,964969}, involving both the statistical modeling and target decomposition methods.

Regarding the statistical modeling, the multiplicative model, which takes into account the contributions of both the backscatter and the speckle, has been suitably employed. 
The return can be modeled by the complex Wishart distribution~\cite{Goodman63,Leegrukwok1994}. 
Other models have been proposed in the literature for PolSAR data, markedly the $\mathcal G_P$ distribution (which has as particular cases the polarimetric $K_P$ and $\mathcal G^0_P$ distributions~\cite{FreitasFrerCorr:2005:PoDiSA,4305361}), the Gaussian scale mixture~\cite{ClassificationNonGaussianPolSAR,ScaleMixturePolSAR}, generalized complex Gaussian laws~\cite{ComplexGeneralizedGaussian}, and the $\mathcal U$ and other models stemming from the multiplicative hypothesis~\cite{OptimalEstimationHeterogeneousPolSAR,SegmentationHeterogeneousPolSAR,
SingleLookMultivariatePolSAR,GoFTestsPolSAR}.
These models are more flexible than the Wishart law (they all include the latter as particular case), at the expense of employing additional parameters whose estimation is oftentimes cumbersome.


Several pixel-based classifiers were developed from the Wishart distribution, being one of them the maximum likelihood classifier used in~\cite{Leegrukwok1994} and the unsupervised procedure employed in~\cite{ImprovingPolSARClassificationCoherencyMatrix}. 
Pixel-based classifiers can be improved by the use of spatial context.
Frery et al.~\cite{4305361} developed an ICM -- Iterative Conditional Modes classifier which employs the maximum likelihood classification result under the complex Wishart distribution as starting point,  point wise evidence, and the Potts model as local information.
This classifier quantifies the spatial information by a maximum pseudolikelihood estimator in a completely manner as segments do.
The Potts model codes the influence of neighboring classes (typically a few, in the implementation here discussed were eight) in a parametric way, whereas a segment is already expected to be a group of data with similar properties.
The ICM algorithm proceeds iteratively until convergence, whereas segment classification by distance minimization is a single-step technique.

It is believed that even better PolSAR classification results can be achieved using segmented images (region-based classification). 
This classification strategy may use a supervised scheme based on stochastic distances between the statistical distributions that model segments and training samples which represent classes. 
In the case of PolSAR data, these distances must be defined between pairs of complex Wishart distributions.

Salicru et al.~\cite{salicru94} developed analytical dissimilarity measures, the so called $\hphi$ family of divergences.
Hypothesis tests based on statistics derived from these divergences were also developed in~\cite{salicru94}. 
Frery et al.~\cite{FreryNascimentoCintraChileanJournalStatistics2011,abraaotestewishart} obtained five  different distances between complex Wishart distributions: Kullback-Leibler, Bhattacharyya, Hellinger, Rényi and Chi-Square and their corresponding hypothesis tests were also developed and evaluated. 

A PolSAR region based classifier using the test statistic derived from the Bhattacharyya stochastic distance between two complex Wishart models was proposed in~\cite{igarss2012}.   The promising results obtained using this classifier in the L band SIR-C image, led us to improve the proposed classifier, by introducing new stochastic distances and their corresponding hypothesis tests.  
In addition to describing in more details the algorithm developed in~\cite{igarss2012},  this article breaks new ground by presenting  analytical expressions for the test statistics based on the following stochastic distances between complex Wishart distributions: Kullback-Leibler, Bhattacharyya, Hellinger, Rényi  and Chi-Square, and also the statistic based on the Bhattacharyya distance between multivariate Gaussian distributions. The classifier performance is evaluated using simulated and real PolSAR data. 
The simulated data is based on the complex Wishart model and the symmetric circularity assumption, aiming at the analysis of such application in statistically well controlled data. 
The real data refer to the complex L-band image, acquired during the 1994 SIR-C mission.

\section{Stochastic Distances and Associated Tests}
\label{sec:stocdist}

Mahalanobis presented the concept of a distance between distributions in the sense that there are pairs of probability laws which are easier to distinguish than others.
Such quantities have received a number of denominations as, for instance, measures of separation, measures of discriminatory information and measures of variation-distance. 
Many goodness-of-fit tests, such as the likelihood ratio, the chi-square, the score and Wald tests, can be defined in terms of appropriate distance measures between distributions.
They all have in common test statistics which increase as the two distributions are further from each other~\cite{llorente2006statistical}.

Salicru et al.~\cite{salicru94} proposed the $\hphi$ family of divergences as follows. 
Consider the random variables $X$ and $Y$ defined on the same support $S$ with distributions characterized by the densities$f_{X}(x;\bm{\theta}_{1})$ and $f_{Y}(x;\bm{\theta}_{2})$, respectively, where $\bm{\theta}_{1}$ and $\bm{\theta}_{2}$ are parameters.
The $\hphi$ divergence between $X$ and $Y$ is given by
\begin{equation} 
\label{eqdivhf}
D^{h}_{\phi}(X,Y)= h\Big(\int_{x\in S}\phi\Big(\frac{f_{X}(x;\bm{\theta}_{1})}{f_{Y}(x;\bm{\theta}_{2})}\Big)f_{Y}(x;\bm{\theta}_{2})dx\Big),
\end{equation}
where $\phi\colon(0,\infty)\rightarrow[0,\infty)$ is a convex function and $h\colon(0,\infty)\rightarrow[0,\infty)$ is a strictly increasing function with $h(0)=0$ and $h'(x)>0$ for all $x\in S$.
Table~\ref{tab-1} presents the choices of $h$ and $\phi$ employed in~\cite{NascimentoCintraFreryIEEETGARS} and the divergences they lead to.

\begin{table*}[hbt]
\centering 
\caption{($h,\phi$)-divergences and related  $\phi$ and~$h$ functions.}\label{tab-1}
\begin{tabular}{rcc}
\toprule
{ $(h,\phi)$-{divergence}} & { $h(y)$} & { $\phi(x)$} \\
\midrule
Kullback-Leibler& $y$ & $x\log(x)$  \\
R\'{e}nyi (order $0<\beta<1$)& $\frac{1}{\beta-1}\log\left((\beta-1)y+1\right),\;0\leq y<\frac{1}{1-\beta}$ & $\frac{x^{\beta}-\beta(x-1)-1}{\beta-1},0<\beta<1$\\
Hellinger  & ${y}/{2},0\leq y<2$ &  $(\sqrt{x}-1)^2$  \\
Bhattacharyya  & $-\log(-y+1),0\leq y<1$ & $-\sqrt{x}+\frac{x+1}{2}$ \\
$\chi^2$ & $y/4$ & $(x-1)^2 (x+1) / x$ \\ \bottomrule
\end{tabular}
\end{table*}

These $\hphi$ divergences are not granted to be symmetric, so they are not necessarily distances.
A simple way to overcome this is computing
\begin{equation}
d_{\phi}^h(X,Y)=\frac{D_{\phi}^h(X,Y)+D_{\phi}^h(Y,X)}{2},
\end{equation}
regardless whether $D_{\phi}^h(\cdot,\cdot)$ is symmetric or not.
Furthermore, if $X$ and $Y$ obey the same distribution with possibly only different parameters, it is enough to write $d_{\phi}^h(\bm{\theta}_{1},\bm{\theta}_{2})$.
Doing so, it is granted that $d_{\phi}^h(\bm{\theta}_{1},\bm{\theta}_{2})=0$ if and only if $\bm{\theta}_{1}=\bm{\theta}_{2}$ and that $d_{\phi}^h(\bm{\theta}_{1},\bm{\theta}_{2}) \geq 0$, but how big this quantity is has no immediate interpretation.

Salicru et al.~\cite{salicru94} provided a means to transform distances into test statistics with known asymptotic properties.
Let $\widehat{\bm{\theta}_{1}}$ and $\widehat{\bm{\theta}_{2}}$ be maximum likelihood estimators of $\bm{\theta}_{1}$ and $\bm{\theta}_{2}$ based on samples of sizes $m$ and $n$, respectively.
The parameter space is $\bm\Theta\subset\mathbbm R^M$.
Under the null hypothesis $H_0:\bm{\theta}_{1}=\bm{\theta}_{2}$, the test statistic
\begin{equation} \label{statistic-1}
S^h_ \phi=\frac{2 mn}{m+n}\frac{d^h_{\phi}(\widehat{\boldsymbol{\theta}_1},\widehat{\boldsymbol{\theta}_2})}{ h{'}(0) \phi{''}(1)} 
\end{equation}
converges in distribution to a $\chi^2_ M$ distributed random variable, where $M$ is the number of parameters of the model, provided $m,n\rightarrow\infty$ such that ${m}/({m+n}) \to \lambda\in(0,1)$.

Nascimento et al.~\cite{NascimentoCintraFreryIEEETGARS} derived $\hphi$ tests for the $\mathcal G^0$ model for intensity SAR data and used them for the discrimination of targets in remote sensing images.
Cintra et al.~\cite{ParametricNonparametricTestsSpeckledImagery} compared those tests with the Kolmogorov-Smirnov test, and verified their robustness.
Frery et al.~\cite{FreryNascimentoCintraChileanJournalStatistics2011} derived these tests for polarimetric SAR data under the Wishart model.
These last results will be recalled in the next section.

\section{Tests Based on Stochastic Distances Between Models}\label{sec:teststoch}

The Wishart law is widely accepted as a model for PolSAR data, mainly on homogeneous areas.
This model stems from the multilook processing of data which obey the complex Gaussian distribution.

We may consider systems with $q$ polarization elements, which record the complex Gaussian distributed random vector $\boldsymbol{y}=(S_1\; S_2\;\cdots\; S_q)^{T}$, where `$T$' denotes vector transposition.
This distribution is characterized by its complex covariance matrix 
$\Sigma=\operatorname{E}(\boldsymbol{y}\boldsymbol{y}^{*}) $, where `$*$' denotes the complex conjugate transpose, and $\operatorname{E}(\cdot)$ is the statistical expectation operator.
In order to enhance the signal-to-noise ratio, $L$ independent and identically distributed samples are usually averaged to form the $L$-looks covariance matrix:	
\begin{equation}
\boldsymbol{Z}=\frac{1}{L}\sum_{i=1}^L \boldsymbol{y}_i\boldsymbol{y}_i^{*}.
\label{eq:L-looks}
\end{equation}
Under these hypotheses, $\boldsymbol{Z}$ follows a scaled complex Wishart distribution with parameters $\boldsymbol{\Sigma}$ and $L$ (denoted by $\boldsymbol{Z}\sim \mathcal W(\boldsymbol{\Sigma},L)$, and characterized by the following probability density function:
\begin{equation}
 f_{\boldsymbol{Z}}({Z};\boldsymbol{\Sigma},L) = \frac{L^{qL}|{Z}|^{L-q}}{|\boldsymbol{\Sigma}|^L \Gamma_q(L)} \exp\bigl(
-L\operatorname{tr}\bigl(\boldsymbol{\Sigma}^{-1}{Z}\bigr)\bigr),
\label{eq:denswishart}
\end{equation}
where $\Gamma_q(L)=\pi^{q(q-1)/2}\prod_{i=0}^{q-1}\Gamma(L-i)$, $L\geq q$, $\Gamma(\cdot)$ is the gamma function, and $\operatorname{tr}(\cdot)$ is the trace operator.
It is important to observe that this Wishart distribution satisfies $\operatorname{E}(\boldsymbol{Z})=\boldsymbol{\Sigma}$. 
The maximum likelihood estimator of $\boldsymbol{\Sigma}$, based on $N$ independent samples, is the sample mean $
\widehat{\boldsymbol{\Sigma}}={N}^{-1}\sum_{i=1}^N {\boldsymbol{Z}_i}$, and $L$ can be estimated by any of the techniques discussed in~\cite{Anfinsen2008}.

Frery et al.~\cite{FreryNascimentoCintraChileanJournalStatistics2011} computed stochastic distances between complex Wishart distributions based on the $\hphi$ divergences presented in Table~\ref{tab-1}, in their most general form (different covariance matrices and number of looks).
In the following we derive the test statistics for the case of same number of looks $L$, assumed known.
The null hypothesis under which these statistics follow a $\chi^2$ distribution is $H_0:\boldsymbol\Sigma_1=\boldsymbol\Sigma_2$.

\begin{align}
S_\text{KL}(\widehat{\boldsymbol{\Sigma}_1},\widehat{\boldsymbol{\Sigma}_2}) &=\frac{2mn}{m+n}L\bigg[\frac{\operatorname{tr}(\widehat{\boldsymbol{\Sigma}_1}^{-1}\widehat{\boldsymbol{\Sigma}_2}+\widehat{\boldsymbol{\Sigma}_2}^{-1} \widehat{\boldsymbol{\Sigma}_1})}{2}-q\bigg].  \label{TestKL}\\
S_\text{B}(\widehat{\boldsymbol{\Sigma}_1},\widehat{\boldsymbol{\Sigma}_2}) &=\frac{8mn}{m+n}L\bigg[\frac{\log|\widehat{\boldsymbol{\Sigma}_1}|+\log|\widehat{\boldsymbol{\Sigma}_2}|}{2}-\log\bigg|\bigg(\frac{\widehat{\boldsymbol{\Sigma}_1}^{-1}+\widehat{\boldsymbol{\Sigma}_2}^{-1}}{2}\bigg)^{-1}\bigg|\bigg]. \label{TestB}\\
S_\text{H}(\widehat{\boldsymbol{\Sigma}_1}, \widehat{\boldsymbol{\Sigma}_2}) &= \frac{8mn}{m+n} \Bigg\{1-\Bigg[\frac{\bigl| 2^{-1} ({\widehat{\boldsymbol{\Sigma}_1}^{-1}+\widehat{\boldsymbol{\Sigma}_2}^{-1}})^{-1} \bigr| }{\sqrt{|\widehat{\boldsymbol{\Sigma}_1}||\widehat{\boldsymbol{\Sigma}_2}|}}\Bigg]^L\Bigg\}. \label{TestH}\\
S_\text{R}^{\beta}(\widehat{\boldsymbol{\Sigma}_1}, \widehat{\boldsymbol{\Sigma}_2}) &= \frac{2mn}{\beta(m+n)}\Bigg\{\frac{\log 2}{1-\beta}+\frac{1}{\beta-1}\log\Big\{\nonumber \\
&\bigl[|\widehat{\boldsymbol{\Sigma}_1}|^{-\beta}|\widehat{\boldsymbol{\Sigma}_2}|^{\beta-1}|(\beta \widehat{\boldsymbol{\Sigma}_1}^{-1}+(1-\beta) \widehat{\boldsymbol{\Sigma}_2}^{-1})^{-1}|\bigr]^L \nonumber\\
&\mbox{}+\bigl[| \widehat{\boldsymbol{\Sigma}_1}|^{\beta-1}| \widehat{\boldsymbol{\Sigma}_2}|^{-\beta}|(\beta \widehat{\boldsymbol{\Sigma}_2}^{-1}+(1-\beta) \widehat{\boldsymbol{\Sigma}_1}^{-1})^{-1}|\bigr]^L\Big\}\Bigg\}. \label{TestR}\\
S_{\chi^2}(\widehat{\boldsymbol{\Sigma}_1}, \widehat{\boldsymbol{\Sigma}_2}) &= \frac{mn}{2(m+n)}\biggl[\biggl(\frac{| \widehat{\boldsymbol{\Sigma}_1}|}{| \widehat{\boldsymbol{\Sigma}_2}|^2}
\operatorname{abs}(|(2 \widehat{\boldsymbol{\Sigma}_2}^{-1}-\widehat{\boldsymbol{\Sigma}_1}^{-1})^{-1}|)\biggr)^L \nonumber\\
& \mbox{}+  \biggl(\frac{|\widehat{\boldsymbol{\Sigma}_2}|}{|\widehat{\boldsymbol{\Sigma}_1}|^2} 
\operatorname{abs}(|(2 \widehat{\boldsymbol{\Sigma}_1}^{-1}-\widehat{\boldsymbol{\Sigma}_2}^{-1})^{-1}|)\biggr)^L -2\biggr],\label{TestChi2}
\end{align}

where `$\operatorname{abs}$' denotes absolute value.
Equations~\eqref{TestKL}, \eqref{TestB}, \eqref{TestH}, \eqref{TestR} and~\eqref{TestChi2} are, respectively, the Kullback-Leibler, Bhattacharyya, Hellinger, R\'enyi of order $\beta$ and $\chi^2$ test statistics based on stochastic distances.
Each test rejects the null hypothesis at level $1-\alpha$ if $\Pr(\chi^2_{q^{2}}\geq s^h_ \phi) \leq \alpha$, where $\chi^2_{q^{2}}$  follows a $\chi^2$ distribution with $q^2$ degrees of freedom.

Notice that Equations~\eqref{TestKL}--\eqref{TestChi2} rely on two simple operations on complex matrices: the inverse and the determinant.

Oftentimes complete PolSAR data are not available.
For instance, Radarsat-2 provides the HH, VV, HV and VH intensities, while dual polarizations are available from
Envisat (HH--HV or VV--VH) and Cosmos Skymed (HH--HV or HH--VV).
In these cases, only elements of the main diagonal of $L$-looks covariance matrix $\boldsymbol Z$ are provided.
Multivariate Gamma models for these data can be derived as marginal distribution from the scaled complex Wishart law characterized by the density given in equation~\eqref{eq:denswishart}.
In practice, such marginal distributions are available for both the bivariate and trivariate cases. The bivariate case, cf.~\cite[Eq.~(30)]{Leehopmanmil94} was used in a maximum likelihood classification algorithm for dual intensity SAR data in~\cite{Leegrupotdual}. Hagedorn et al.~\cite{Hagedorn:2006} derived both the bi- and tri-variate $\chi^{2}$ distributions of diagonal elements of a Wishart law, but there are currently no expressions available for the distances between these multivariate chi-squared distributions.

Additionally, an increased number of looks and the amplitude format yield a distribution which can be approximated by a multivariate Gaussian law.
This, and the fact that multivariate Gaussian classifiers are a commodity of image processing software, suggests the use of the Gaussian model as a testbed for the data here considered.

Theodoridis and Koutroumbas~\cite{PRT} compute stochastic distances under the $q$-variate Gaussian model.
Using these results and  equation~\eqref{statistic-1}, we derived the Bhattacharyya test statistic for the null hypothesis $H_0:(\boldsymbol{\mu}_1,\boldsymbol{\Sigma}_1)=(\boldsymbol{\mu}_2,\boldsymbol{\Sigma}_2)$:
\begin{align}
T_\text{B} &= \frac{8mn}{m+n}\Bigg[ [(\widehat{\boldsymbol{\mu}_{1}}-\widehat{\boldsymbol{\mu}_{2}})^{T} \Big(\frac{\widehat{ \boldsymbol{\Sigma}_1}+\widehat{\boldsymbol{\Sigma}_2}}{2}\Big)^{-1} (\widehat{\boldsymbol{\mu}_{1}}-\widehat{\boldsymbol{\mu}_{2}})]+ 4  \log \frac{ \big|\frac{\widehat{\boldsymbol{\Sigma}_1} + \widehat{\boldsymbol{\Sigma}_2}}{2}\big|}{\sqrt{|\widehat{\boldsymbol{\Sigma}_1}| |\widehat{\boldsymbol{\Sigma}_2}|}}  \Bigg] \label{batgaussdist},
\end{align}
where $\widehat{\boldsymbol{\mu}_{i}}$ and $\widehat{\boldsymbol{\Sigma}_i}$ are the maximum likelihood estimators of the mean vector and the covariance matrix, $i=1,2$. The null hypothesis is rejected at level $1-\alpha$ if $\Pr(\chi^2_{{q(q+3)}/{2}}\geq T_\text{B}) \leq \alpha$, where $\chi^2_{{q(q+3)}/{2}}$  follows a $\chi^2$ distribution with ${q(q+3)}/{2}$ degrees of freedom.

\section{Region Classification based on Test Statistics}\label{sec:Classification}

In this section we define the two classification products we obtain using test statistics based on stochastic distances: mininum test statistics and $p$-value maps.

Assume the image support is partitioned in $r$ disjoint segments $C_1,\dots,C_r$.
The PolSAR data from each segment is denoted $Z_{C_{i}}$, $1\leq i\leq r$, and a covariance matrix $\widehat{\boldsymbol\Sigma_i}$ is estimated with these data by maximum likelihood.
The user provides $k$ prototypes in the form of samples (supervised scheme), with which covariance matrices $\widehat{\boldsymbol\Sigma_\ell}$, $1\leq \ell\leq k$, are estimated by maximum likelihood.
The purpose is to classify each segment $C_i$ in one of the $k$ prototypes.

Compute the $r\times k$ test statistics which contrast the null hypothesis $H_0: \boldsymbol{\Sigma}_i = \boldsymbol{\Sigma}_\ell$ with one of the equations given in~\eqref{TestKL}--\eqref{TestChi2} for every segment $1\leq i\leq r$ and every prototype $1\leq \ell\leq k$.
The classification based on minimum test statistic consists of assigning the segment $C_i$ to the class represented by prototype $t$ if
\begin{equation}
S^h_\phi(\widehat{\boldsymbol{\Sigma}_i} , \widehat{\boldsymbol{\Sigma}_t})  <
S^h_\phi(\widehat{\boldsymbol{\Sigma}_i} , \widehat{\boldsymbol{\Sigma}_\ell}) \label{eq:ClassificationRule}
\end{equation}
for every $t\neq \ell$.
Once the segment $C_i$ has been assigned to the class represented by prototype $t$, the $p$-value of the assigment is computed as
\begin{equation}
p_{i,t} = \Pr(\chi^2_{\nu} > s^h_\phi(\widehat{\boldsymbol{\Sigma}_i} , \widehat{\boldsymbol{\Sigma}_t})), \label{eq:Pvalue}
\end{equation}
where $\nu$ is the numbers of parameters of the considered model: $\nu = q^2$  for the Wishart distribution, and $\nu = {q(q+3)}/{2}$ for the $q$-variate Gaussian distribution. 
This value gives an idea of the confidence of the decision.

The rule given by inequality~\eqref{eq:ClassificationRule} opens a number of interesting alternatives, among them, instead of choosing \textbf{one} test statistic, use all available ones. Each test statistic will provide a class for each segment, and these classifications can be fused by majority vote.
The information provided by equation~\eqref{eq:Pvalue} can also be used; a fuzzy classification can be made for each segment to \textbf{all} the classes whose $p$-value is above a certain threshold.

\section{Application to PolSAR Data}

The classification procedure described in Section~\ref{sec:Classification} was applied and evaluated under two approaches: using simulated data, which was generated under the complex Wishart distribution, and using a real SIR-C full PolSAR image, in L-band.

\subsection{SIR-C Polarimetric Data Description}

\begin{figure}[hhh]
\centering
\subfigure[Training samples \label{trainn}]{\includegraphics[width=6cm]{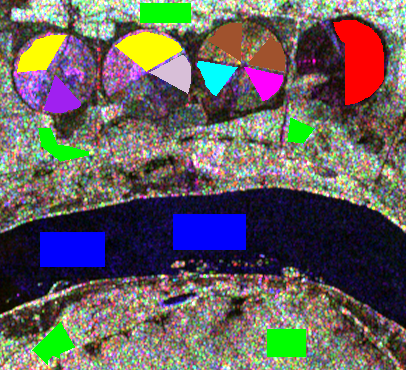}}
\subfigure[Test samples \label{test}]{\includegraphics[width=6cm]{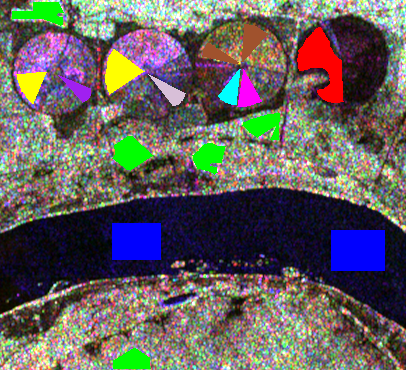}}\\
\subfigure[Land cover classes legend \label{leg}]{\includegraphics[width=8cm]{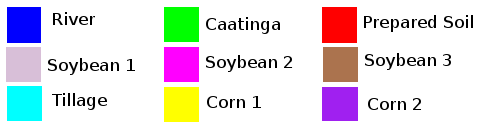}}\\
\caption{L-band SIR-C intensity  images (HH(R), HV(G), VV(B)) and location of training and test samples.}\label{Vresults}
\end{figure}

The SIR-C full polarimetric image is from an agricultural area located in Petrolina city, Northeast of Brazil. 
Table~\ref{tab:ImageInformation} presents the study area location and the basic characteristics of the SIR-C image. 
The main observed land cover classes  are River, Caatinga, Prepared Soil, Soybean in three different phenological  stages, Tillage, and Corn in two phenological stages. The training and test samples for these classes are shown in Figures~\ref{trainn} and \ref{test}, and their legends in Figure~\ref{leg}. These samples were properly sub-sampled  to diminish the pixels correlation influence, and their final sizes are shown in Table \ref{tab:traintestnumber}.

\begin{table}[hbt]
\caption{SIR-C image and study area information.}\label{tab:ImageInformation}
\centering
\begin{tabular}{l p{5cm}}
\toprule
Study area location & $09\degree\ 07\arcminute$~S, $40\degree\ 18\arcminute$~W (central coordinate), about \unit[$40$]{\kilo\metre} northeast of the city of Petrolina-PE, Brazil \\
Aquisition date &  April $14^{th}$, 1994 \\
Image size (pixels) &  $407\times 370$ \\
Nominal number of looks & 4.785\\
Frequency &  L-band - \unit[$1.254$]{GHz} \\
Pixel spacing & $12.5 m \times 12.5 m$\\
Incidence angle & 49.496\degree\\
Orbit direction & Descending\\
\bottomrule
\end{tabular}
\end{table}

\begin{table*}[ht]
\caption{training and test samples description.}\label{tab:traintestnumber}
\centering
\begin{tabular}{l p{7cm} c c}
\toprule
Class & Description &\# Training samples & \# Test samples \\
\hline
River & Water body &$1192$ & $976$\\
Caatinga & A stepped vegetation composed of stunted trees and thorny bushes, found in areas of little rainfall in Brazil &$1006$ & $820$\\
Prepared Soil & Soil ready for seeding  & $715$ & $442$\\
Soybean 1 & Soybean with approximately 52 days after seeding&$212$ & $99$\\
Soybean 2 & Soybean with approximately 66 days after seeding&$174$ & $117$\\
Soybean 3 & Soybean with approximately 113 days after seeding&$390$ & $216$\\
Tillage & Agricultural crops residuals &$181$ & $98$\\
Corn 1 & Corn with less than 124 days after seeding&$661$ & $364$\\
Corn 2 & Corn with approximately 133 days after seeding&$191$ & $77$\\
\bottomrule
\end{tabular}
\end{table*}

Frery et al.~\cite{4305361} concluded that, with the exception of the class ``River'',  the samples presented in Fig.~\ref{Vresults} depart from the  Wishart distribution~\cite[page.~7, Table~III]{4305361} and are better explained by the $\mathcal{K}_P$ and $\mathcal{G}^{0}_{P}$ distributions.
As previously mentioned, there are no analytic expressions for the stochastic distances between such generalized models, and numerical integration would be unfeasible due the need to integrate on the domain of all positive definite Hermitian matrices.
In this manner, although the exact description of the data could be improved, adopting the Wishart model still leads to interesting results.

\subsection{Simulated Data Description}\label{simulation}

Simulated data were generated under the symetric circularity assumption~\cite{Goodman63}. 
The simulation aims at obtaining random covariance matrices realizations under the complex Wishart distribution with a fixed number of looks ($L$). 
Initially, single-look polarimetric SAR data, represented by the $q$-variate complex Gaussian random  vector $\boldsymbol{y}_q$, are generated. 
Assuming that $\boldsymbol{y}_q$ follows a $q$-variate complex Gaussian distribution with zero mean and covariance matrix $\boldsymbol{\Sigma}_q$ (denoted $\boldsymbol{y}_q\sim \mathcal{CN}_{q}(0,\boldsymbol{\Sigma}_{q})$), the simulation is performed by first sampling a $2q$-variate vector $\boldsymbol{x}$ such that $\boldsymbol{x}_{2q}\sim \mathcal{N}_{2q}(0,\boldsymbol{\Sigma}_{2q}^{*})$, where, under the symetric circular assumption and according to~\cite{Goodman63} and \cite{Picinbono96}, $\boldsymbol{\Sigma}_{2q}^{*}$ is such that:
 \begin{equation*}
  \boldsymbol{\Sigma}_{2q}^{*} = \frac{1}{2} \left[ \begin{array}{cc}\Re(\boldsymbol{\Sigma}_q) & -\Im(\boldsymbol{\Sigma}_q) \\ \Im(\boldsymbol{\Sigma}_q) & \Re(\boldsymbol{\Sigma}_q)\end{array} \right],
 \end{equation*}
 where $\Re$ and $\Im$ denote the real and imaginary parts of a complex number, respectively.
The first $q$ elements of $\boldsymbol{x}_{2q}$ become the real parts of the elements in the complex vector $\boldsymbol{y}_q$ and the last $q$ elements of $\boldsymbol{x}_{2q}$ become the imaginary parts of the elements in the complex vector $\boldsymbol{y}_q$.
This process is repeated as many times as the required number of samples, where each sample represents a polarimetric pixel of an image.
The $L$-looks complex covariance matrix image is obtained according to equation~\eqref{eq:L-looks}. 

\begin{figure}[hbt]
\centering
\subfigure[\label{simul}]{\includegraphics[width=6cm]{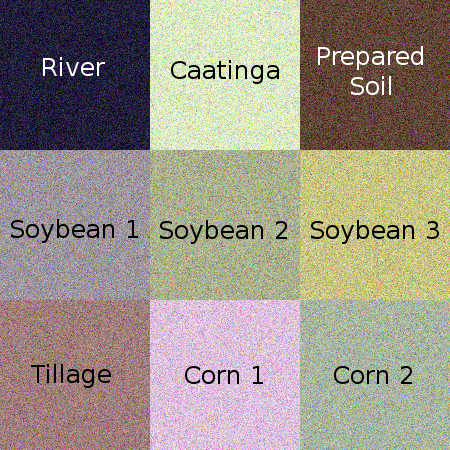}}\quad
\subfigure[\label{seg15}]{\includegraphics[width=6cm]{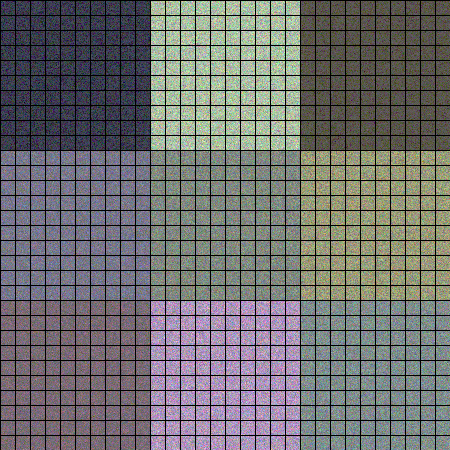}}\\
\caption{Simulated PolSAR image: (a)  intensities color composition - HH(R), HV(G), VV(B)  and (b) segmentation scheme in $15\times15$ pixels segments.}\label{cmsimul}
\end{figure}

The simulation process described above was used to produce images representing  the nine classes observed in the SIR-C L-band PolSAR image. 
The covariance matrices for each class were the estimated covariance matrices, using the training samples presented in Figure~\ref{trainn},  whose numbers of pixels are shown in Table~\ref{tab:traintestnumber}. 
These  covariance matrices are presented in equations~\eqref{riv}--\eqref{c2} of the appendix.  
The simulation was performed with four looks and three polarization bands, HH, HV and VV. 
The simulated covariance matrix image of each class has $150\times150$ pixels. 
A final image was generated by mosaicking the simulated images of the individual classes. 
This final image  has $450\times450$ pixels, i.e., the images were grouped in a $3\times3$ images classes configuration. 
An RGB color composition of the intensities bands from the covariance matrix image is shown in Figure~\ref{cmsimul}.

The region classification procedure was applied using four different segmentation procedures to all the segments of  sizes $5\times5$, $10\times10$, $15\times15$ and $30\times30$ pixels, respectively. 
The  $15\times15$ segmented image is presented in Figure~\ref{seg15}. 
The prototype of each class, also needed for the classification procedure, was generated by sampling $900$ pixels, representing a training sample of $30\times30$ pixels for each class.
The simulation of the prototypes, performed independently of the  simulated image, ensures that identical data are not being considered in the computation of test statistics and, consequently, in the determination of the corresponding $p$-values​​.

\subsection{Assessing the Classification Procedure using Simulated Data}

The region-based classifications of the simulated data, using stochastic distances and the minimum test statistics given in equation~\eqref{eq:ClassificationRule}, aimed at the evaluation of the classification procedure under rigorous well controlled statistical model, as the data was simulated considering the complex Wishart distribution. 
An additional classification result was obtained, considering the multivariate Gaussian model for the multivariate amplitude image obtained from the simulated PolSAR image. 
For this case, the analytic  expression for the Bhattacharyya test statistic showed in equation~\eqref{batgaussdist} was used. 

Observing that we have four segmented images and six stochastic distances (five from Wishart  and one from Gaussian models), twenty four classified images were produced.
The classifications performed on segments of sizes $10\times10$, $15\times15$ and $30\times30$ pixels were $100\%$ correct. 
The classifications of segments of size $5\times5$ pixels reached a global accuracy of $99.81\%$ for the Bhattacharyya, Kullback-Leibler, Hellinger and Rényi distances, and $99.58\%$ for the $\chi^2$ distance. 
Errors occurred in segments belonging to the simulated classes of Soybean~2 and Corn~2. 

These results show the high quality of the proposed classifier when the assumptions of the data distribution are satisfied, especially for segments with large amounts of pixels (equal or greater than $100$ pixels).  

\begin{figure}[hbt]
\centering
\subfigure[Bhattacharyya \label{bt}]{\includegraphics[width=4cm]{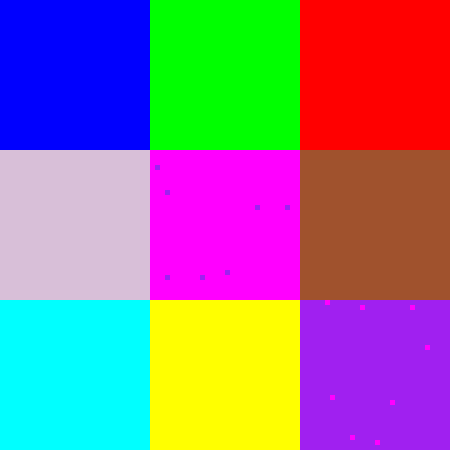}}\quad
\subfigure[Kullback-Leibler \label{kl}]{\includegraphics[width=4cm]{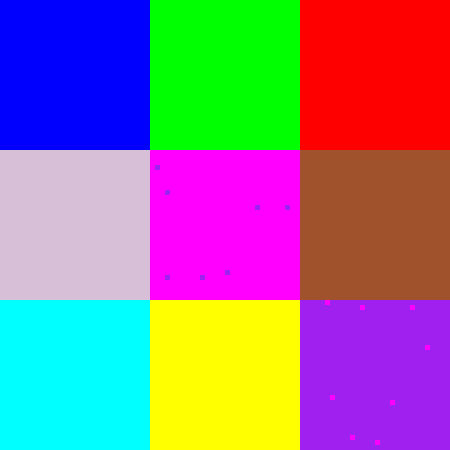}}\quad
\subfigure[Hellinger  \label{h}]{\includegraphics[width=4cm]{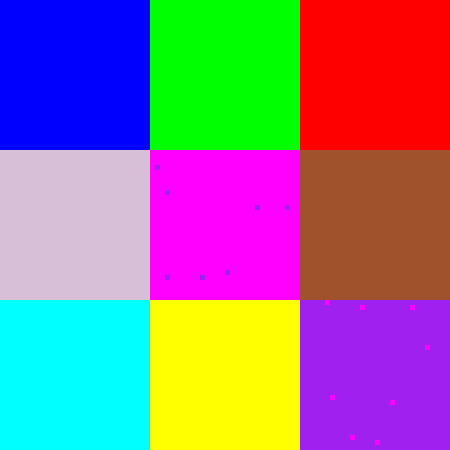}}\quad
\subfigure[{\scriptsize Rényi of order $\beta$}  \label{r09}]{\includegraphics[width=4cm]{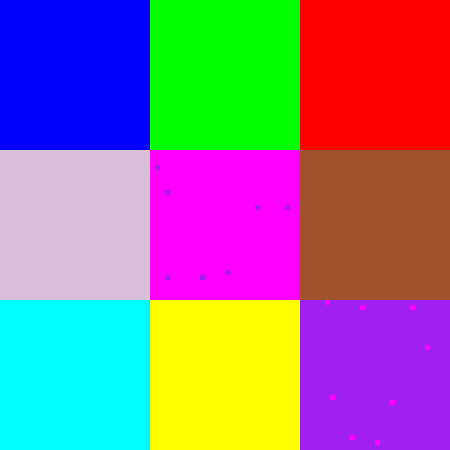}}\quad
\subfigure[$\chi^{2}$  \label{chi2}]{\includegraphics[width=4cm]{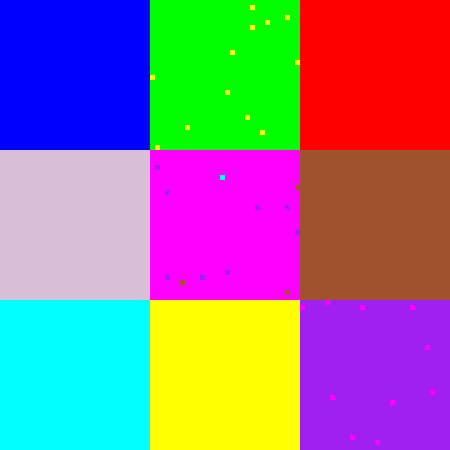}}\quad
\subfigure[Bhattacharyya Gaussian \label{btg}]{\includegraphics[width=4cm]{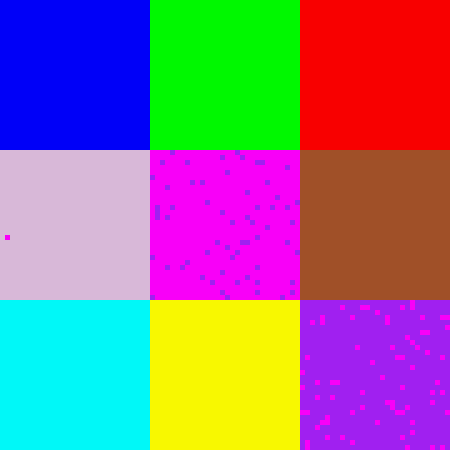}}
\subfigure{\includegraphics[width=7cm]{./figs/legenda.png}}\\
\caption{Classification results of the simulated data for segments of size $5\times5$ pixels.}\label{classsimul5}
\end{figure}

Figure~\ref{classsimul5} shows the classified images using the six stochastic distances, for the case of  $5\times5$ pixels segments. 
The slight confusion between the Soybean~2 and Corn~2 classes can be observed in this figure, as well as with some segments of the Caatinga class, classified as Corn~1 class under the $\chi^2$ distance. 
Under the Gaussian model, the global accuracy was $98.35\%$ for the Bhattacharyya distance. 
A higher confusion between  Soybean~2 and Corn~2 classes was observed when compared with the results obtained by the classifiers that adopt the Wishart model, a result which stresses the importance of using the proper distribution to model the data.

\begin{figure*}[hbt]
\centering
\mbox{
\subfigure[\scriptsize Bhattacharyya $5\times5$ \label{}]{\includegraphics[width=3.0cm]{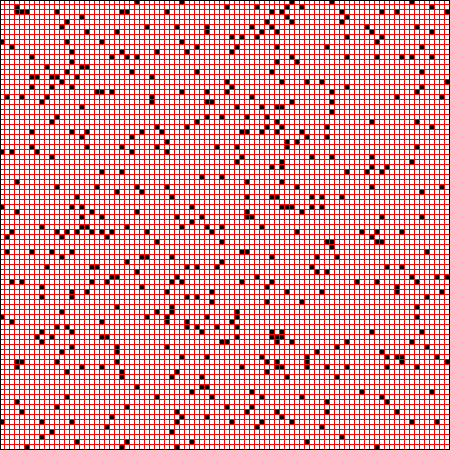}}\quad
\subfigure[\scriptsize Bhattacharyya $10\times10$ \label{}]{\includegraphics[width=3.0cm]{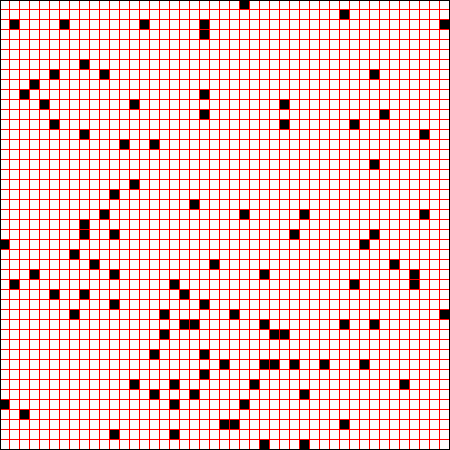}}\quad
\subfigure[\scriptsize Bhattacharyya $15\times15$  \label{}]{\includegraphics[width=3.0cm]{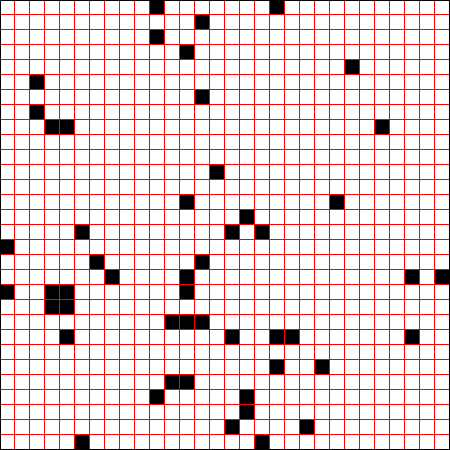}}\quad
\subfigure[\scriptsize Bhattacharyya $30\times30$  \label{}]{\includegraphics[width=3.0cm]{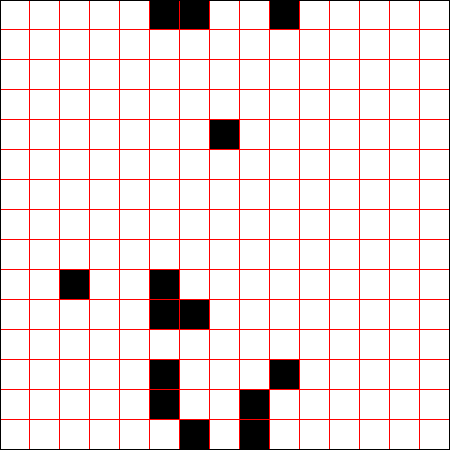}}
}

\mbox{
\subfigure[\scriptsize Kullback-Leibler $5\times5$ \label{}]{\includegraphics[width=3.0cm]{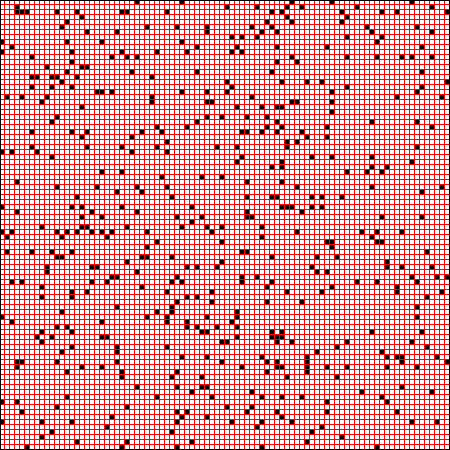}}\quad
\subfigure[\scriptsize Kullback-Leibler $10\times10$ \label{}]{\includegraphics[width=3.0cm]{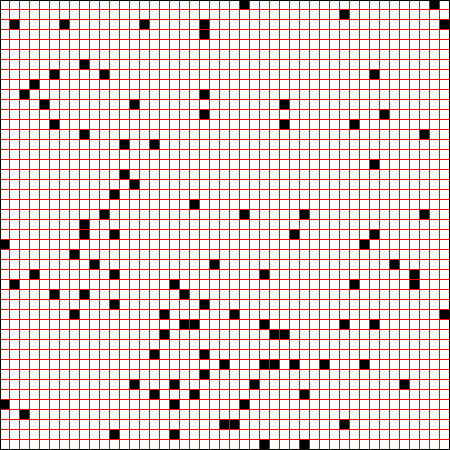}}\quad
\subfigure[\scriptsize Kullback-Leibler $15\times15$  \label{}]{\includegraphics[width=3.0cm]{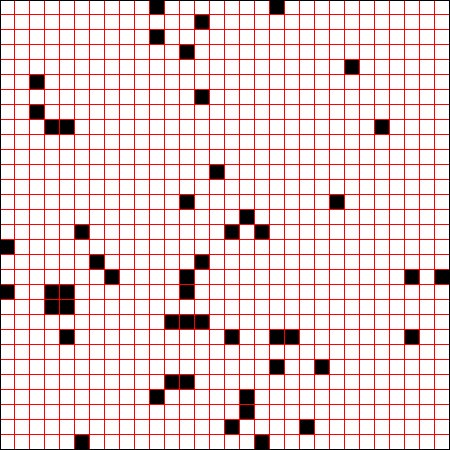}}\quad
\subfigure[\scriptsize Kullback-Leibler $30\times30$  \label{}]{\includegraphics[width=3.0cm]{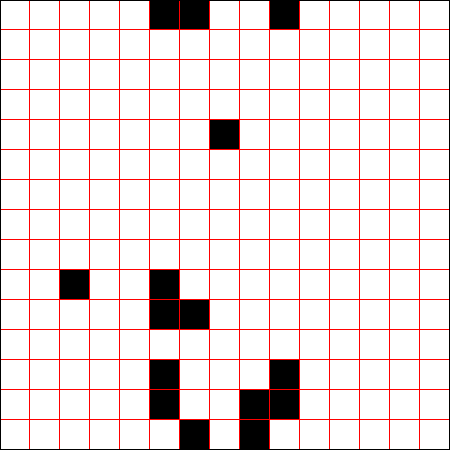}}
}

\mbox{
\subfigure[\scriptsize Hellinger $5\times5$ \label{}]{\includegraphics[width=3.0cm]{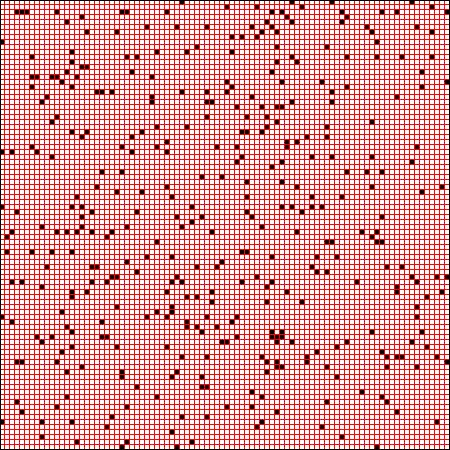}}\quad
\subfigure[\scriptsize Hellinger $10\times10$ \label{}]{\includegraphics[width=3.0cm]{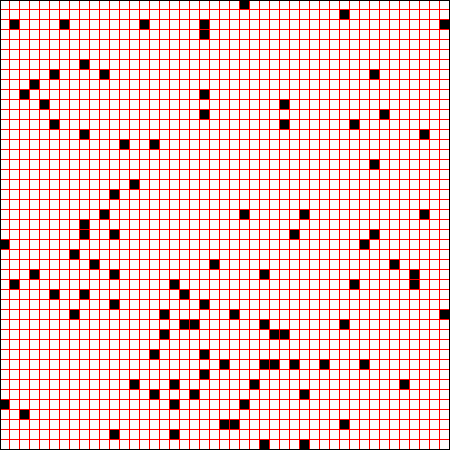}}\quad
\subfigure[\scriptsize Hellinger $15\times15$  \label{}]{\includegraphics[width=3.0cm]{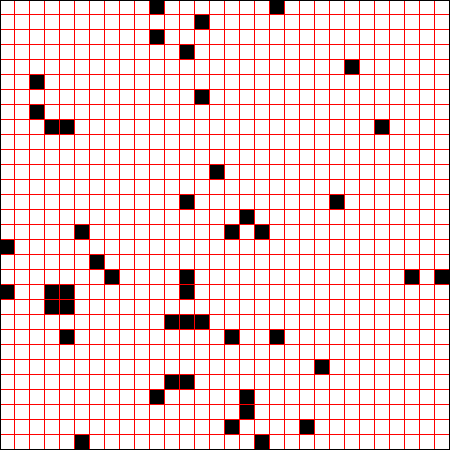}}\quad
\subfigure[\scriptsize Hellinger $30\times30$  \label{}]{\includegraphics[width=3.0cm]{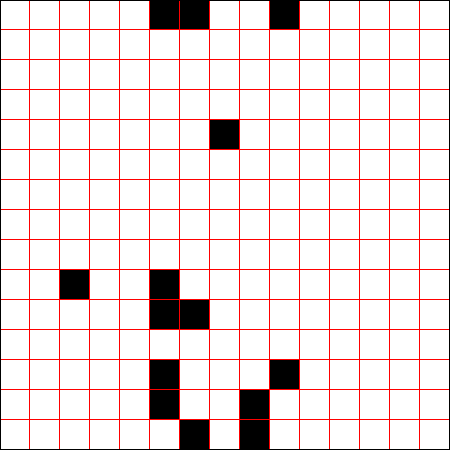}}
}

\mbox{
\subfigure[{\scriptsize Rényi of order $\beta$ $5\times5$} \label{}]{\includegraphics[width=3.0cm]{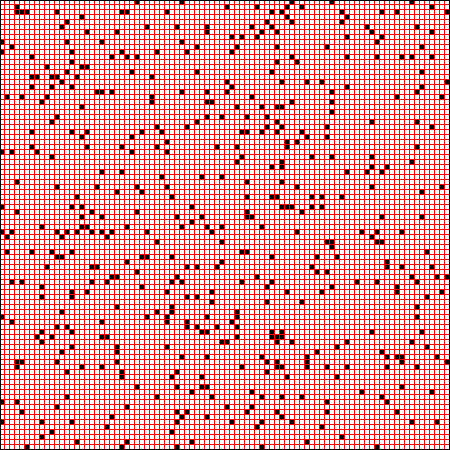}}\quad
\subfigure[{\scriptsize Rényi of order $\beta$ $10\times10$} \label{}]{\includegraphics[width=3.0cm]{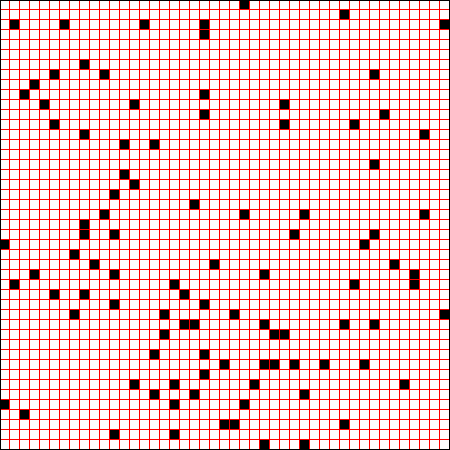}}\quad
\subfigure[{\scriptsize Rényi of order $\beta$ $15\times15$} \label{}]{\includegraphics[width=3.0cm]{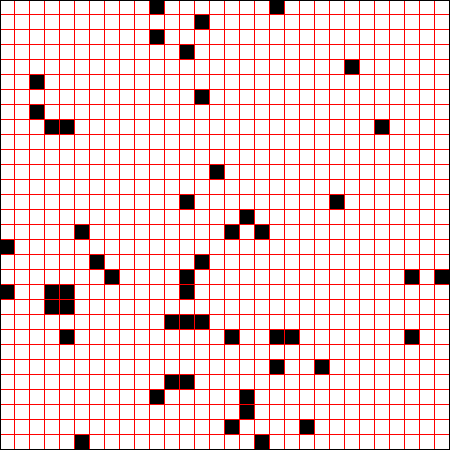}}\quad
\subfigure[{\scriptsize Rényi of order $\beta$ $30\times30$}  \label{}]{\includegraphics[width=3.0cm]{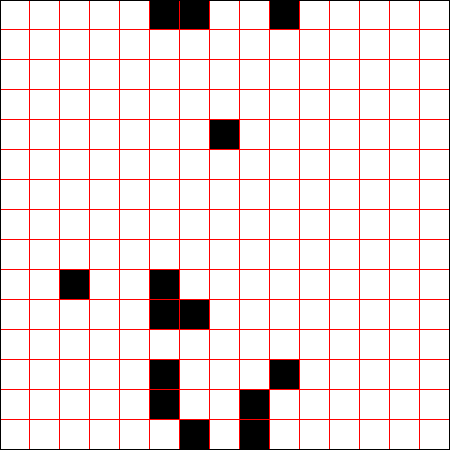}}
}

\mbox{
\subfigure[\scriptsize $\chi^{2}$  $5\times5$ \label{}]{\includegraphics[width=3.0cm]{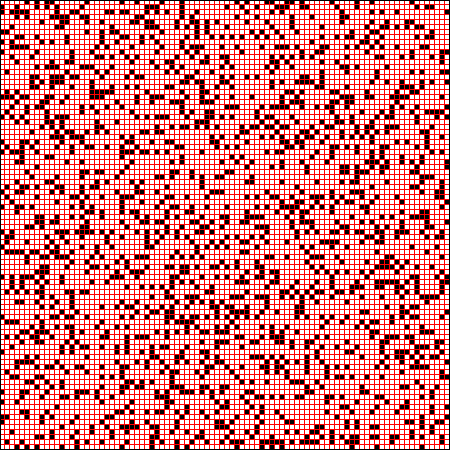}}\quad
\subfigure[\scriptsize $\chi^{2}$  $10\times10$ \label{}]{\includegraphics[width=3.0cm]{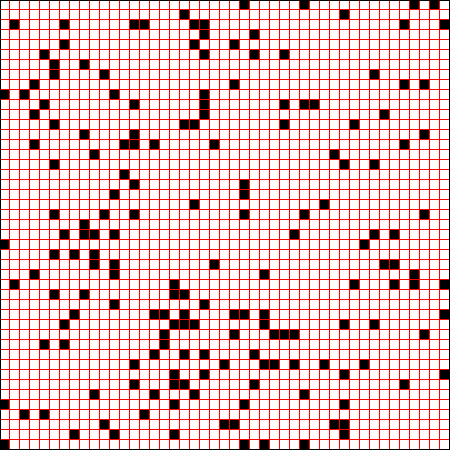}}\quad
\subfigure[\scriptsize $\chi^{2}$  $15\times15$  \label{}]{\includegraphics[width=3.0cm]{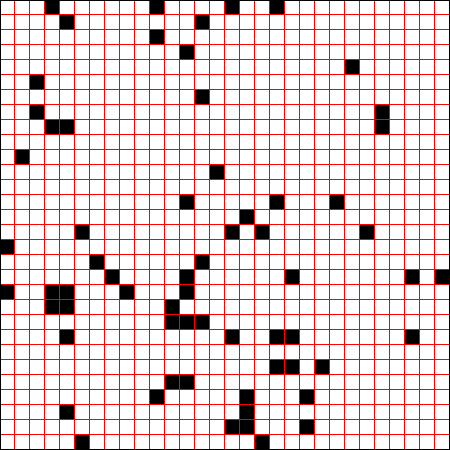}}\quad
\subfigure[\scriptsize $\chi^{2}$  $30\times30$  \label{}]{\includegraphics[width=3.0cm]{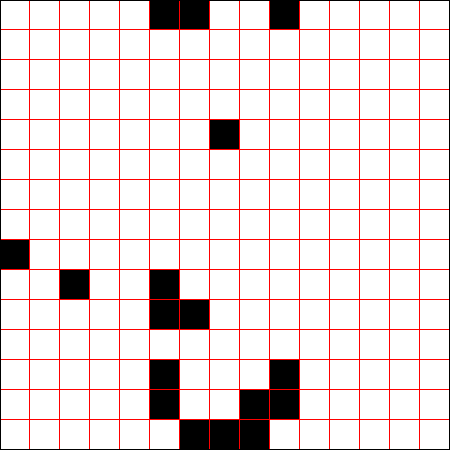}}
}

\mbox{
\subfigure[\scriptsize Bhatt. Gaussian $5\times5$ \label{}]{\includegraphics[width=3.0cm]{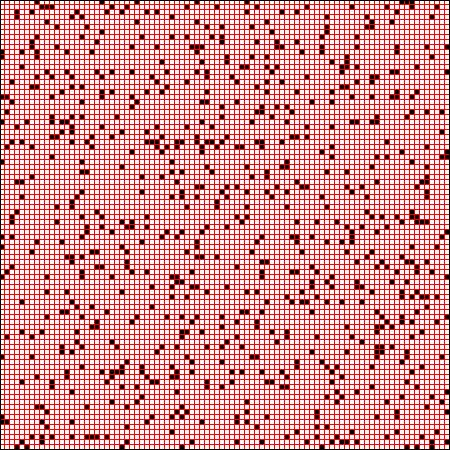}}\quad
\subfigure[\scriptsize Bhatt. Gaussian $10\times10$ \label{}]{\includegraphics[width=3.0cm]{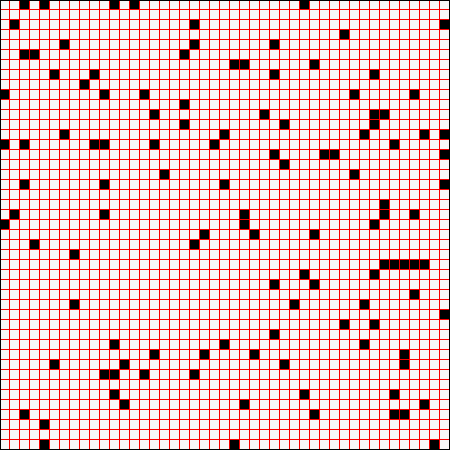}}\quad
\subfigure[\scriptsize Bhatt. Gaussian $15\times15$  \label{}]{\includegraphics[width=3.0cm]{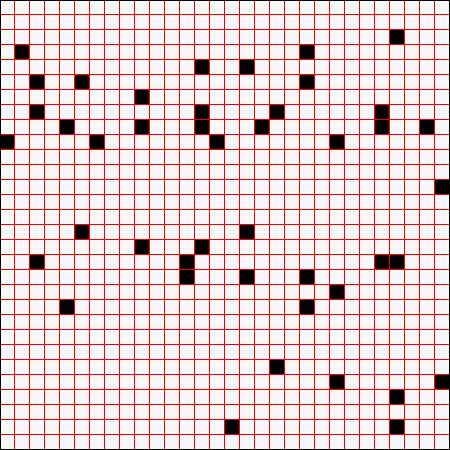}}\quad
\subfigure[\scriptsize Bhatt. Gaussian $30\times30$  \label{}]{\includegraphics[width=3.0cm]{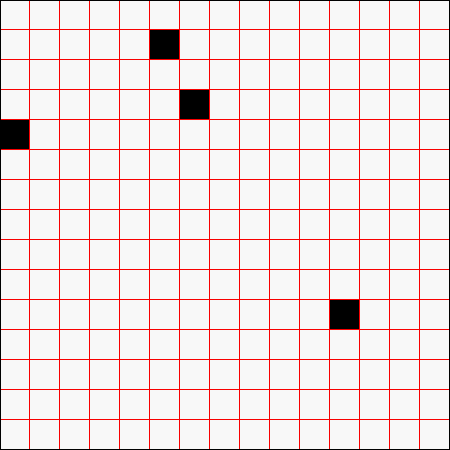}}
}
\caption{$P$-value binary maps for the classified simulated images - segments in white have  null hypothesis accepted ($p \geq 0.05$).}\label{resulspsimul}
\end{figure*}

For each classification result, a map of the $p$-values of the test statistics, an indicator of the confidence of the assignment decision, was also produced.
These results are presented in Figure~\ref{resulspsimul}, where the white positions mark those segments for which the null hypothesis (the equality between the covariance matrices of the segment and of the assigned prototype)  was not rejected at the \unit[$5$]{\%} significance level. 
The percentages of these segments for each sample size and each distance is presented in Table~\ref{segpercsimul}.

\begin{table*}[htbp]
\caption{Percentage of segments for which $H_0$ was not rejected at  \unit[$5$]{\%} significance level, for  simulated data case.}\label{segpercsimul}
\centering
\begin{tabular}{c c c c c}
\toprule
\multirow{3}{*}{Distances} & \multicolumn{4}{c}{Percentage (\%)}\\
\cmidrule(lr){2-5}
&$5\times5$ pixels&$10\times10$ pixels&$15\times15$ pixels&$30\times30$ pixels\\
&\unit[$8100$]segments&\unit[$2025$] segments&\unit[$900$] segments&\unit[$225$]segments\\
\midrule
Bhattacharrya&$94.0$&$95.2$&$94.3$&$93.8$\\
Kullback-Leibler&$93.7$&$95.1$&$94.3$&$93.3$\\
Hellinger&$95.2$&$95.3$&$94.8$&$93.8$\\
Rényi (order $\beta=0.9$)&$93.8$&$95.1$&$94.3$&$93.8$\\
$\chi^{2}$&$75.5$&$91.2$&$92.8$&$92.4$\\
Bhattacharrya (Gaussian)&$90.6$&$94.1$&$95.1$&$98.2$\\
\bottomrule                                                                                                                                                                                                                        
\end{tabular}
\end{table*}

The results presented in  Figure~\ref{resulspsimul} and in  Table~\ref{segpercsimul} are ​​compatible  with the theoretically expected values. 
The hypothesis tests rejection rates were approximately $5\%$ for all  segmentation cases and stochastic distances, except when  the  $\chi^{2}$ distance was used, and the Bhattacharrya Gaussian distance was applied to small  ($5\times5$ pixels) segments. 
The rejection rates for the $\chi^{2}$ distance were higher than the theoretical values in all segmentation cases, reaching the value of approximately \unit[$24.5\%$] for the segmentation of $5\times5$ pixels segments. 
The poor performance of the $\chi^2$ distance test statistic was also observed by \cite{FreryNascimentoCintraChileanJournalStatistics2011}, where this big test size was first described. 

The rejection rate of the Hellinger distance is $5\%$ in $5 \times 5$, $10 \times 10$ and $15 \times 15$ pixels segments, while the Bhattacharrya Gaussian distance  reaches this rate only in large segments ($15 \times 15$ and $30 \times 30$), as expected according to the Central Limit Theorem.

\subsection{Assessing the Classification Procedure using SIR-C Polarimetric Image}

Prior to classification, the SIR-C image was segmented using the SegSAR software~\cite{SousaJr2005}: a hierarchical multi-level region growing segmentation algorithm designed for intensity SAR data which uses tests based on the Gamma and Gaussian distributions.
The SegSAR parameters used for segmentation were \unit[$100$]{pixels} of minimum area, and \unit[$1$]{dB} of similarity. 

\begin{sidewaysfigure*}
\centering
\subfigure[Segmented image \label{fig:SegmentedLband}]{\includegraphics[width=.24\linewidth]{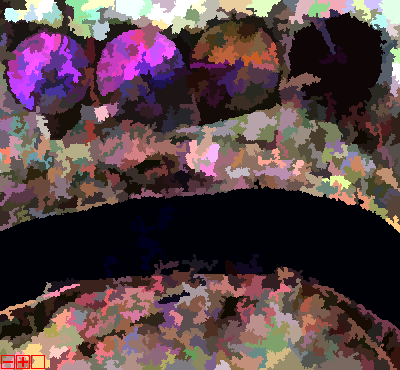}}
\subfigure[Bhattacharyya]{\includegraphics[width=.24\linewidth]{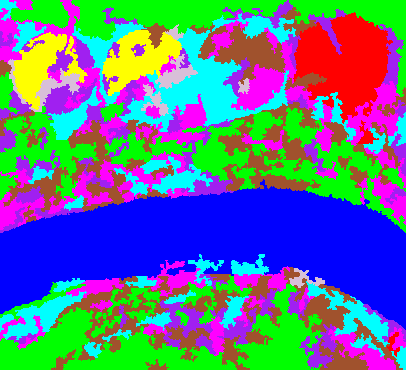} \label{Lband_bat}}
\subfigure[Kullback-Leibler]{\includegraphics[width=.24\linewidth]{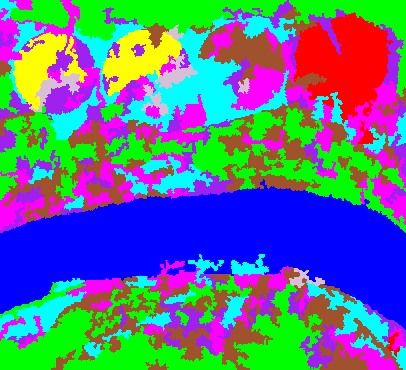} \label{Lband_kl}}
\subfigure[Hellinger]{\includegraphics[width=.24\linewidth]{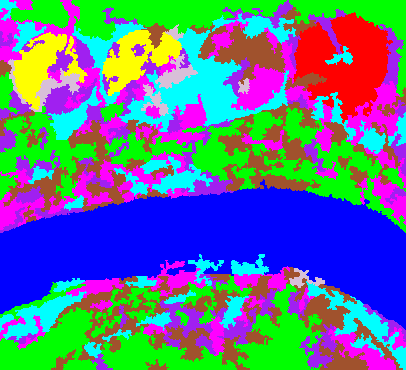} \label{Lband_hel}}
\subfigure[Rényi of order $\beta=0.9$]{\includegraphics[width=.24\linewidth]{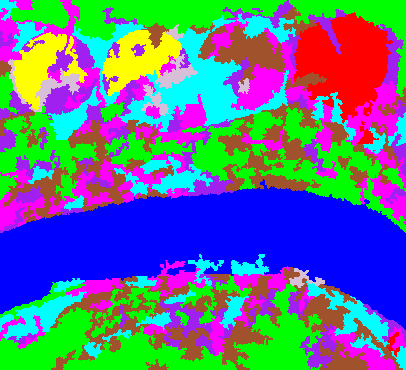} \label{Lband_re}}
\subfigure[$\chi^{2}$]{\includegraphics[width=.24\linewidth]{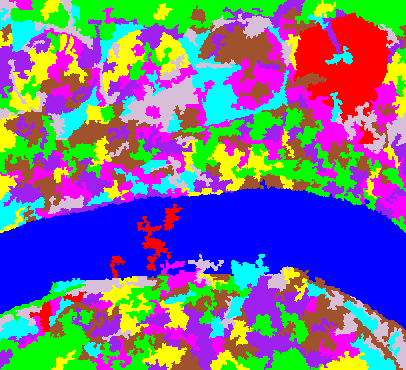} \label{Lband_chi}}
\subfigure[Bhattacharyya Gaussian]{\includegraphics[width=.24\linewidth]{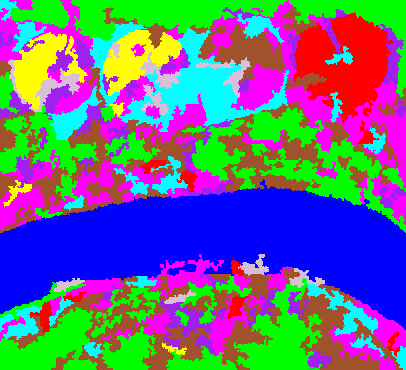} \label{Lband_batg}}
\subfigure[Contextual ICM]{\includegraphics[width=.24\linewidth]{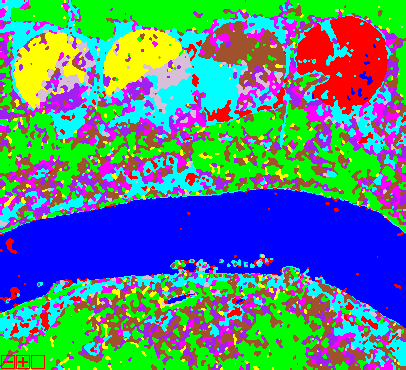} \label{Lband_ICM}}
\caption{SIR-C image  segmentation and classification results.}\label{Lbandclassification}
\end{sidewaysfigure*}

The equivalent number of looks value was estimated considering all polarization channels using the method described in~\cite{4305361}, which is also referred by Anfisen et al.~\cite{Anfinsen2008} as Fractional Moment Estimate; the computed value was $2.97$. 
The segmented image is presented in Figure~\ref{fig:SegmentedLband}; each segment is shown in a color defined by associating the RGB channels to the means of each intensity polarization (HH, HV and VV).
The classification procedure described by equation~\eqref{eq:ClassificationRule} was applied for L-band SIR-C data using this segmented image.
The tests statistics are given in equations~\eqref{TestKL}--\eqref{TestChi2}, assuming the Wishart law, as well as equation~\eqref{batgaussdist}, assuming the Gaussian law for amplitude data. 
The $p$-value map, defined in equation~\eqref{eq:Pvalue}, was computed for every segment in each classification.
 
The region classifications were also compared to the contextual ICM polarimetric classification described in~\cite{4305361}, which is also based on the Wishart distribution. 
Therefore,  six region classifications and one ICM classification were obtained. 
The classification performances were compared using the estimated Kappa coefficient of agreement ($\hat{\kappa}$), and the overall accuracy, as formulated in~\cite{Congalton2009Assessing}.

The classifications results are presented in Figure~\ref{Lbandclassification}. 
The overall accuracy, the estimated Kappa coefficient of agreement and its variance for all classifications are presented in Table~\ref{kappa}.  
The tests for equality of Kappa showed that the classifications based on  Kullback-Leibler, Bhattacharrya, Hellinger and Rényi distances between Wishart distributions produced statistically similar results.  
The contextual classification is only superior to the $\chi^2$ distance classification, which is the worst stochastic distance-based classification, in agreement  with the results found with simulated data.
The classification based on the Bhattacharrya Gaussian distance is only superior to the contextual and to the  $\chi^2$ distance classification. 

\begin{sidewaysfigure*}
\centering
\subfigure[Bhattacharyya]{\includegraphics[width=.24\linewidth]{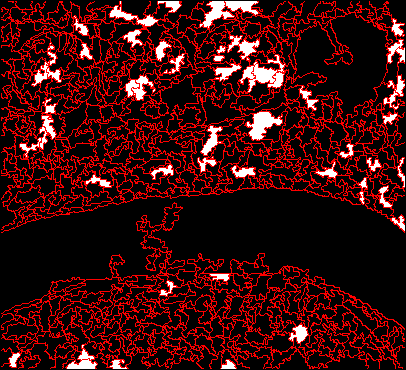} \label{Lband_bat_sp}}
\subfigure[Kullback-Leibler]{\includegraphics[width=.24\linewidth]{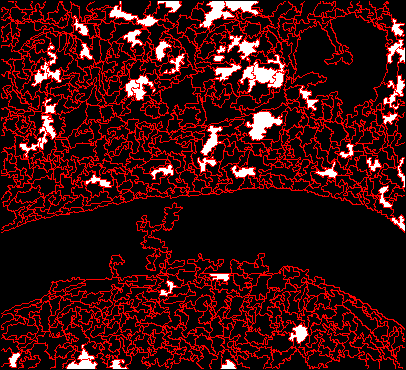} \label{Lband_kl_sp}}
\subfigure[Hellinger]{\includegraphics[width=.24\linewidth]{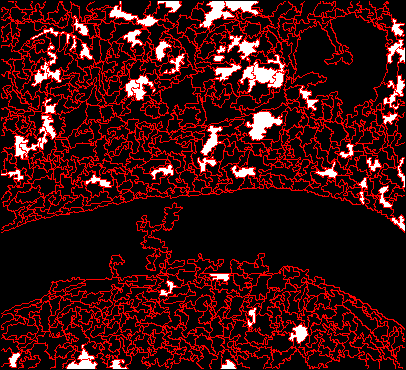} \label{Lband_hel_sp}}
\subfigure[Rényi of order $\beta=0.9$]{\includegraphics[width=.24\linewidth]{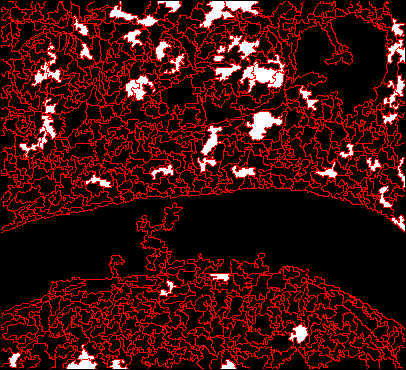} \label{Lband_re_sp}}
\subfigure[$\chi^{2}$]{\includegraphics[width=.24\linewidth]{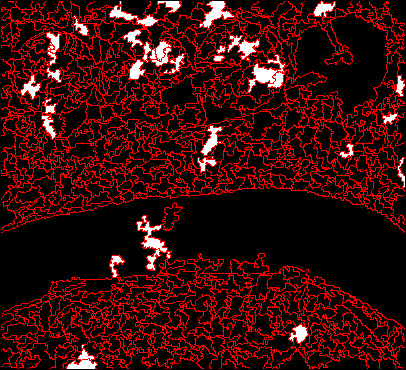} \label{Lband_chi_sp}}
\subfigure[Bhattacharyya Gaussian]{\includegraphics[width=.24\linewidth]{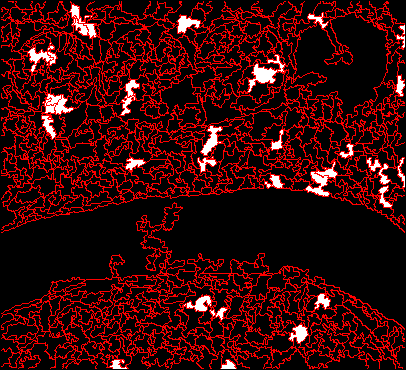} \label{Lband_batg_sp}}
\caption{$P$-value binary maps for SIR-C image classifications - segments in white have  null hypothesis accepted ($p \geq 0.05$.)}\label{Lbandsegmap}
\end{sidewaysfigure*}

Table~\ref{segperc} presents, for each stochastic distance, the percentage of segments with $p$-value greater than $0.05$, i.e., the percentage of segments that were not rejected at this level.  
These segments are illustrated in white in Figure~\ref{Lbandsegmap}.

Although the values of Table~\ref{kappa} showed promising results for minimum distance classification using complex Wishart distributions, the percentages showed in Table~\ref{segperc} are far from the  theoretical $95\%$. 

\begin{table*}[hbt]
\centering
\caption{Assessment of classification results for L-band SIR-C image.}\label{kappa}
\setlength{\tabcolsep}{1.7pt}
\begin{tabular}{c|cccc}
\toprule
 \multicolumn{2}{c}{Classification Method}  & \multicolumn{3}{c}{L-band SIR-C Image}\\
\cmidrule(lr){3-5} 
 \multicolumn{2}{c}{ } & Overall Accuracy (\%)&$\hat{\kappa}$&$s_{\hat{\kappa}}(\times10^{-5})$\\
\midrule
\multirow{6}{*}{\rotatebox{90}{Region-based}} 
&Bhattacharrya&$86.60$&$0.8346$&$1.253$\\
&Kullback-Leibler&$86.60$&$0.8346$&$1.253$\\
&Hellinger&$85.97$&$0.8269$&$1.296$\\
&Rényi (order $\beta=0.9$)&$86.60$&$0.8346$&$1.253$\\
& $\chi^{2}$&$71.36$&$0.6544$&$2.081$\\
&Bhattacharrya (Gaussian)&$85.35$&$0.8191$&$1.333$\\
\midrule
Contextual &ML/ICM - Wishart&$83.97$&$0.8025$&$1.430$\\
\bottomrule        
\end{tabular}
\end{table*}

\begin{table*}[htbp]
\caption{Percentage of segments of the SIR-C L-band image that were not rejected at the \unit[$5$]{\%} significance level.}\label{segperc}
\centering
\begin{tabular}{c r}
\toprule
Distances & Percentage (\%)\\
\midrule
Bhattacharrya&$9.76$\\
Kullback-Leibler&$9.58$\\
Hellinger&$10.49$\\
Rényi (order $\beta=0.9$)&$9.58$\\
$\chi^{2}$&$6.33$\\
Bhattacharrya (Gaussian)&$6.33$\\
\bottomrule                                                                                                                                                                                                                        
\end{tabular}
\end{table*}

\section{Conclusions and future work}

A new region-based classifier for PolSAR data using stochastic distances between complex Wishart distributions and derived hypothesis tests was presented.
The proposed classifier was applied to simulated data and to a real L-band PolSAR image from the SIR-C mission.

The classification results using simulated PolSAR data based on the complex Wishart model obtained an overall acccuracy of $100\%$, with the exception of few misclassification observed when small segments ($5\times5$ pixels) were used. 
The acceptance rates of the null hypothesis tests, which measures the confidence of the classification assignments, obtained very close values to the theoretically expected ones for almost all distances and segment sizes. 
The poorest results occurred when the $\chi^{2}$ distance was used, especially in the classifications of small segments. 
With the exception of this last case, the evidence allows us to conclude that the proposed classification method has a very good performance and confidence when the data rigorously follow the Wishart model.
Further evaluations with non-perfect Wishart independent observations, such as under presence of noise, departures from the pure model, and spatial correlation are under implementation and investigation.

The use of statistic based on the Hellinger distance between Wishart laws usually outperforms the results obtained by other distances, specially for segmented images with small regions.
This may be due to the robustness which the procedures derived from this distance have.
The Battacharrya Gaussian distance is also a good option for images having large segments.

The proposed region-based classifier, when applied to L-band PolSAR data from the SIR-C mission, obtained also very good performance in terms of overall accuracy and $\kappa$ coefficient of agreement. 
The best results were obtained with the Bhattacharrya, Kullback-Leibler, Rényi and Hellinger distances between Wishart distributions. 
The results using these distances overcame the classification results obtained using multivariate amplitude data and the Bhattacharrya distance between Gaussian laws. 
This evidence proves the relevance of using appropriate modeling of the data when employing stochastic distances.

In comparison with a contextual Maximum Likelihood/ICM classifier~\cite{4305361}, the new classifier obtained also  better results, with statistically superior $\kappa$ values.  
Such improvement can also be observed by examining the huge amount of undesirable small areas that still exist in the contextual result, while those artifacts are minimized by the region-based classification. 

The rejection rates of the null hypothesis tests concerning the real PolSAR data was distant from the theoretical expected values, achieving values higher than \unit[$90\%$]. 
Since the results with complex Wishart simulated data were perfectly compatible with the theoretical expected significance level, this poor result with real data may be due to a less than optimal description of the real data by the theoretical model. 
As mentioned before, many samples are better modeled by more general distributions as the $\mathcal{K}_P$  and $\mathcal{G}^{0}_{P}$ laws. 
These results suggest that the proposed method is robust with respect to the classification map, but not the map of $p$-values.

Another possible sources of misfit are spatial correlation, which alters the effective sample sizes in Eq.~\eqref{statistic-1}, the existence of more classes than those identified by the expert, the large size of some segments (as in the ``River'' class), and the influence of an inadequate segmentation (the SegSAR algorithm used in this paper was developed for intensity and not for PolSAR data). 
Further investigation must be taken forward with real data examples in order to clarify the possible vulnerability of hypothesis testing due to these possibilities. 

The analysis of the proposed region-based classification led us to conclude that the classifier has great potential for PolSAR data analysis.
It is noteworthy that the expressions that have to be computed rely on simple operations on matrices: the determinant and the inverse.
In the future, further investigation will be conducted using also the classifier module considering the intensity pair distribution for bivariate intensity data, a common SAR data availability situation commented in Section~\ref{sec:teststoch}.

Recent research~\cite{MultivariateTextureRetrievalGeodesicDistance,GeodesicsTextureDiscrimination} reports interesting results with the use of the Geodesic Rao metric~\cite{GeodesicEstimationEllipticalDistributions}.
This, and other tests statistics for hypothesis testing PolSAR data distributions, along with improved~\cite{VasconcellosFrerySilva:CompStat,SilvaCribariFrery:ImprovedLikelihood:Environmetrics,CribariFrerySilva:CSDA} and robust~\cite{AllendeFreryetal:JSCS:05,BustosFreryLucini:Mestimators:2001} estimation in models which incorporate texture~\cite{FreitasFrerCorr:2005:PoDiSA,PolarimetricSegmentationBSplinesMSSP} are future lines of research.

\appendix

The covariance matrices of the nine classes of the SIR-C images were estimated by maximum likelihood using the selected training samples (Figure~\ref{trainn} and Table~\ref{tab:traintestnumber}).  
Equations (\ref{riv}) to (\ref{c2}) present the estimated covariance matrices for the following classes: River, Caatinga, Prepared Soil, Soybean~1, Soybean~2, Soybean~3, Tillage, Corn~1, and Corn~2, respectively. 
These matrices were the parameter used for image simulation under the Wishart model, as described in section~\ref{simulation}. Only the upper triangle and the diagonal are displayed in the equations (\ref{riv}) to (\ref{c2}) because the covariance matrix ($\boldsymbol{\Sigma}$) is Hermitian and, therefore, the remaining elements are
the complex conjugates.

\begin{figure*}
\hrulefill
\begin{align}
\boldsymbol{\Sigma}_{\text{river}} = \left[\begin{array}{ccc} 2.98\cdot10^{-3} & 5.31\cdot10^{-6} + \jmath 8.11\cdot10^{-5} & 3.47\cdot10^{-3} + \jmath 3.42\cdot10^{-4} \\  & 3.40\cdot10^{-4} & 4.47\cdot10^{-6} + \jmath 1.39\cdot10^{-4} \\  &  & 1.19\cdot10^{-2} \end{array} \right] \label{riv}\\
\boldsymbol{\Sigma}_{\text{caatinga}} = \left[\begin{array}{ccc} 1.11\cdot10^{-1} & -3.10\cdot10^{-3} - \jmath 1.58\cdot10^{-3} & 1.98\cdot10^{-2} + \jmath 1.65\cdot10^{-3} \\  & 3.40\cdot10^{-2} & -1.41\cdot10^{-3} + \jmath 1.87\cdot10^{-3} \\ &  & 9.47\cdot10^{-2} \end{array} \right] \label{caa}\\
\boldsymbol{\Sigma}_{\text{prep soil}} = \left[\begin{array}{ccc} 1.05\cdot10^{-2} & -5.39\cdot10^{-6} - \jmath 2.37\cdot10^{-4} & 7.53\cdot10^{-3} + \jmath 1.75\cdot10^{-3} \\  & 8.46\cdot10^{-4} & -3.38\cdot10^{-5} + \jmath 1.32\cdot10^{-4} \\  &  & 1.14\cdot10^{-2} \end{array} \right] \label{ps}\\
\boldsymbol{\Sigma}_{\text{soybean 1}} = \left[\begin{array}{ccc} 3.40\cdot10^{-2} & -1.79\cdot10^{-3} - \jmath 1.86\cdot10^{-3} & -3.6\cdot10^{-4} - \jmath 7.58\cdot10^{-3} \\  & 5.16\cdot10^{-3} & 4.38\cdot10^{-4} + \jmath 4.28\cdot10^{-4} \\ &  & 5.38\cdot10^{-2} \end{array} \right] \label{s1}\\
\boldsymbol{\Sigma}_{\text{soybean 2}} = \left[\begin{array}{ccc} 4.31\cdot10^{-2} & -1.76\cdot10^{-3} - \jmath 1.32\cdot10^{-3} & -1.78\cdot10^{-4} - \jmath 1,73\cdot10^{-3} \\ & 9.26\cdot10^{-3} & 6.55\cdot10^{-4} + \jmath 1.27\cdot10^{-3} \\ & & 4.35\cdot10^{-2} \end{array} \right] \label{s2}\\
\boldsymbol{\Sigma}_{\text{soybean 3}} = \left[\begin{array}{ccc} 7.53\cdot10^{-2} & -4.25\cdot10^{-3} - \jmath 7.66\cdot10^{-3} & 5.87\cdot10^{-4} - \jmath 1.36\cdot10^{-3} \\  & 1.47\cdot10^{-2} & -2.18\cdot10^{-4} + \jmath 1.21\cdot10^{-3} \\ \cdot&  & 3.70\cdot10^{-2} \end{array} \right] \label{s3}\\
\boldsymbol{\Sigma}_{\text{tillage}} = \left[\begin{array}{ccc} 3.53\cdot10^{-2} & 1.20\cdot10^{-3} + \jmath 1.02\cdot10^{-4} & 1.64\cdot10^{-2} - \jmath 2.65\cdot10^{-3} \\ & 3.05\cdot10^{-3} & 4.48\cdot10^{-4} + \jmath 1.88\cdot10^{-4} \\  & & 3.29\cdot10^{-2} \end{array} \right] \label{tl}\\
\boldsymbol{\Sigma}_{\text{corn 1}} = \left[\begin{array}{ccc} 1.15\cdot10^{-1} & -3.95\cdot10^{-3} - \jmath 3.57\cdot10^{-3} & 9.13\cdot10^{-3} - \jmath 4.86\cdot10^{-3} \\ & 1.33\cdot10^{-2} & 3.34\cdot10^{-3} + \jmath 2.83\cdot10^{-3} \\  &  & 1.47\cdot10^{-1} \end{array} \right] \label{c1}\\
\boldsymbol{\Sigma}_{\text{corn 2}} = \left[\begin{array}{ccc} 4.19\cdot10^{-2} & 1.08\cdot10^{-3} - \jmath 1.01\cdot10^{-3} & 9.24\cdot10^{-3} - \jmath 3.68\cdot10^{-3} \\  & 1.02\cdot10^{-2} & 2.43\cdot10^{-4} + \jmath 3.31\cdot10^{-4} \\  &  & 5.71\cdot10^{-2} \end{array} \right] \label{c2}
\end{align}
\hrulefill
\end{figure*}


\bibliographystyle{IEEEbib}
\bibliography{./refs}

\begin{IEEEbiography}[{\includegraphics[width=1in,height=1.25in,clip,keepaspectratio]{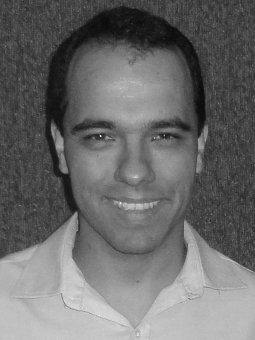}}]{Wagner B. Silva}
received the B.Sc. and M.Sc. degrees in cartographic engineering from the Instituto Militar de Engenharia, Rio de Janeiro, Brazil, in 1999 and 2005, respectively. He is currently working toward the Ph.D. degree in Remote Sensing in the Instituto Nacional de Pesquisas Espaciais (INPE), São José dos Campos, Brazil. His research interests include stochastic models and SAR image processing.
\end{IEEEbiography}

\begin{IEEEbiography}[{\includegraphics[width=1in,height=1.25in,clip,keepaspectratio]{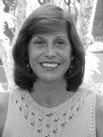}}]{Corina C. Freitas}
received the B.S. degree in mathematics from the Pontifícia Universidade Católica de São Paulo in 1974, the M.Sc. degree in statistics from the Massachusetts Institute of Technology, Cambridge, in 1980, and the Ph.D. degree in statistics from the University of Sheffield, UK, in 1992. She is currently a Researcher at the Instituto Nacional de Pesquisas Espaciais (INPE), São José dos Campos, Brazil. Her research interests include statistical analysis of SAR images and SAR image processing.
\end{IEEEbiography}

\begin{IEEEbiography}[{\includegraphics[width=1in,height=1.25in,clip,keepaspectratio]{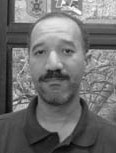}}]{Sidnei J. S. Sant’Anna}
received the B.S. degree in electrical and electronic engineer from the Universidade Federal do Rio de Janeiro in 1993,
the M.Sc. degree in Remote Sensing from the Instituto Nacional de Pesquisas Espaciais (INPE), São José dos Campos, Brazil in 1995 and the Ph.D. degree in Eletronic Engineering and Computing from the Instituto Tecnológico de Aeronáutica (ITA), São José dos Campos, Brazil, in 2009. He is currently a researcher at INPE, and his interests are image analysis and processing techniques for remote sensing (SAR image filtering, statistical methods, robustness, etc.).

\end{IEEEbiography}

\begin{IEEEbiography}[{\includegraphics[width=1in,height=1.25in,clip,keepaspectratio]{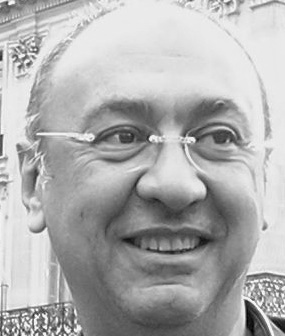}}]{Alejandro C. Frery (S’92–M’95)}
received the B.S. degree in electronic and electrical engineering from the Universidad de Mendoza, Mendoza, Argentina, the M.Sc. degree in applied mathematics (statistics) from the Instituto de Matemática Pura e Aplicada, Rio de Janeiro, Brazil, and the Ph.D. degree in applied computing from the Instituto Nacional de Pesquisas Espaciais, São José dos Campos, Brazil. He is currently with the Instituto de Computação, Universidade Federal de Alagoas, Maceió, Brazil. His research interests are statistical computing and stochastic modeling.
\end{IEEEbiography}

\end{document}